\documentclass[letterpaper, 10 pt, conference]{ieeeconf}  

\usepackage{graphicx} 
\usepackage{amsmath} 
\usepackage{spconf}
\title{\bf
Multi-atlas based pathological stratification of d-TGA congenital heart disease
}

\name{M.A. Zuluaga$^{1}$, A. Mendelson$^{1}$, M.J. Cardoso$^{1}$, A.M. Taylor$^{2,3}$, S. Ourselin$^{1}$}
\address{$^{1}$ Centre for Medical Image Computing (CMIC), University College London, UK\\
$^{2}$ Centre for Cardiovascular Imaging, UCL Institute of Cardiovascular Science, London, UK\\
$^{3}$ Cardiorespiratory Division, Great Ormond Street Hospital for Children, London, UK}

\begin{document}

\maketitle
\thispagestyle{empty}
\pagestyle{empty}

\begin{abstract}

One of the main sources of error in multi-atlas segmentation propagation approaches comes from the use of atlas databases that are morphologically dissimilar to the target image.  In this work, we exploit the segmentation errors associated with poor atlas selection to build a computer-aided diagnosis (CAD) system for pathological classification in post-operative dextro-transposition of the great arteries (d-TGA). The proposed approach extracts a set of features, which describe the quality of a segmentation, and  introduces them into a logical decision tree that provides the final diagnosis. We have validated our method on a set of 60 whole heart MR images containing healthy cases and two different forms of post-operative d-TGA. The reported overall CAD system accuracy was of 93.33\%.

\end{abstract}

\section{INTRODUCTION}
Dextro-transposition of the great arteries (d-TGA) is a congenital heart disease in which the two major vessels that carry blood away from the heart, \textit{i.e.} the aorta and the pulmonary artery, are switched. Although  d-TGA is repaired at birth through two different procedures known as arterial and atrial switch, monitoring of the patient during their life time is crucial. Identifying the anatomical variations of treated d-TGA from MR images is a complex task that requires high clinical expertise and that can be missed during screening. Therefore, an accurate system capable of diagnosing the condition is highly desirable.

To date, there has been little use of computer-aided diagnosis (CAD) in cardiovascular diseases. Previous works~\cite{duchateau,zuluaga2011} have attempted to define normality and to identify deviating patterns as a pathology (abnormality). Our aim is to go further by being capable to point abnormalities and specify the underlying pathology. For this matter, we make use of multi-atlas segmentation propagation (MASP).

\begin{figure}[t]
\begin{center}
\includegraphics[width=0.77\columnwidth]{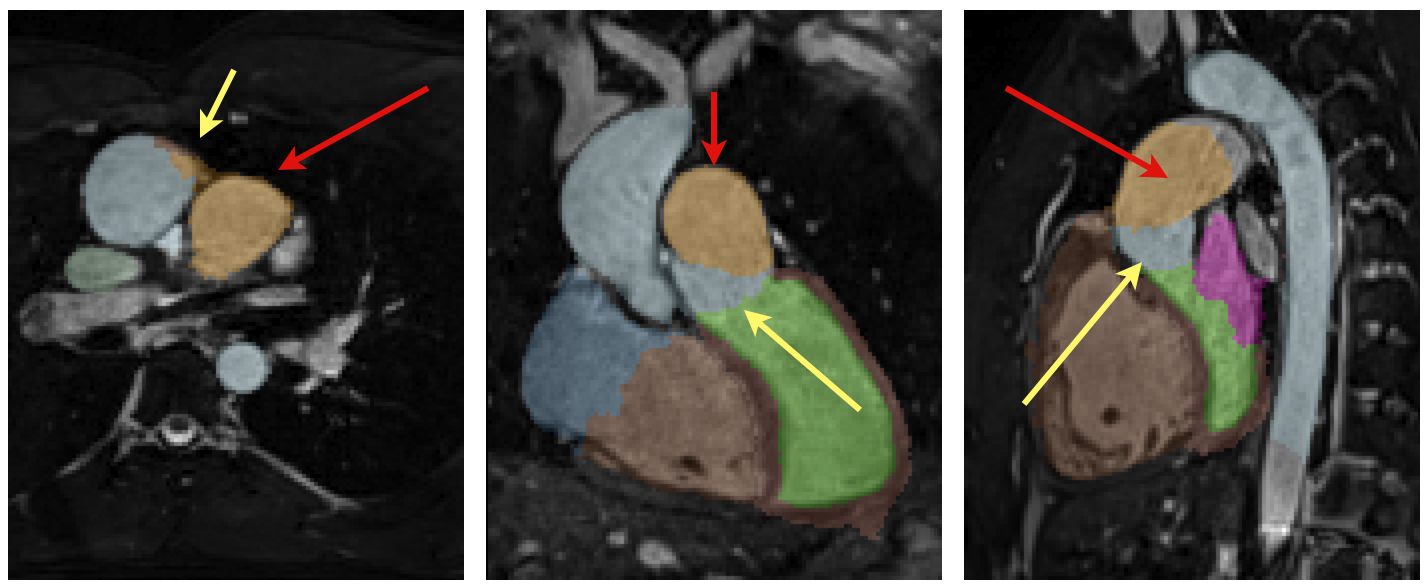} 
\end{center}
\caption{3D MR whole heart scan presenting atrial switch d-TGA and the (overlaid) segmentation obtained using a healthy atlas. Red arrows point to correctly segmented portions of the pulmonary artery (orange). Yellow arrows show segmentation errors caused by the lack of morphological similarity between the target and the atlas set.}
\label{fig:fusionerrors}
\end{figure}

MASP is a well established method for segmenting images where a database of annotated atlases is available~\cite{cardoso2012}. An image segmentation is obtained by transforming the set of atlases into the unseen or target image space and then applying a fusion method to combine the label images from each atlas into a consensus segmentation. In order to obtain a good segmentation, there must be sufficient morphological similarity between the atlases and the target image. When this condition is not satisfied, the existence of a diffeomorphic  transformation between the atlases and the target image space cannot be guaranteed, leading to unrepresentative segmentations (Fig.~\ref{fig:fusionerrors}). This situation is common when pathologies introduce morphological alterations in the affected organs. 

As the quality of a segmentation is dependent of the morphological difference between a target image and an atlas set, it forms a natural source of information for the classification of a disease in an automatic or semi-automatic manner. Thus, a MASP knowledge based system can help distinguish between non homeomorphic pathologies. In this work, we introduce a CAD system based on a logical, binary decision tree which combines segmentation quality measures into a final diagnosis of post-operatory d-TGA.

\section{METHOD}
The proposed method builds over the hypothesis that well and poorly segmented images have different distributions in some representative feature space, making it possible to discriminate them. As the use of an unrepresentative atlas is more likely to lead to a poor segmentation, this discrimination tells us about the similarity of a newly segmented image to the atlas set used. Thus, when the images in an atlas are known to belong to a morphological subgroup, we can decide whether or not a patient shares the same underlying condition. We shall denote the set of well segmented images as the \textit{positive} or similar class, and the set of poorly segmented images as belonging to the \textit{negative} or dissimilar class. 

In the simple scenario with only two possible outcomes, \textit{e.g} healthy vs. pathological, the class discrimination is easily applied for diagnosis. The unseen image is segmented by propagating the labels of a healthy atlas database followed by the extraction of a feature set that allows classification of the segmentation into the \textit{positive} or \textit{negative} class. When more clinical conditions are introduced for diagnosis, it is necessary to introduce a scheme that combines the simple \textit{positive}/\textit{negative} classification into a higher level diagnosis system. We will refer to this decision scheme as a logical decision tree. In the next sections, we first provide further details on how \textit{positive} vs. \textit{negative} class discrimination is achieved, followed by an introduction to the logical decision tree concept. 

\subsection{Positive vs. Negative Class Discrimination}
Given $A$, an atlas database of images sharing a specific morphological/pathological pattern, the goal of the \textit{positive} vs. \textit{negative} class discrimination is to determine if a segmented image is morphologically similar to $A$ or not. We propose a three step solution: 
multi-atlas segmentation propagation (MASP), feature set extraction, and classification. 

\textbf{Multi-atlas Segmentation Propagation.}
The MASP step  is performed by using a whole heart segmentation framework that combines a two-stage registration algorithm with a multi-label fusion approach as described in~\cite{zuluaga2013}. To determine if an unseen image contains a specific pathology, it has to be segmented using an atlas set with the same pathological pattern. 

\textbf{Feature Set Extraction.}
Our feature set definition is inspired by the processes underlying human visual perception, described by the so-called Gestalt perception laws~\cite{desolneux2003}. 
Whenever a point (or a higher order visual object) shares one or more characteristics with other points, it will get grouped with other points to form a new higher order visual object denoted as a \textit{Gestalt}. Examples of grouping characteristics include proximity, intensity similarity and continuity, among others. A correctly segmented anatomical structure falls into the definition of a \textit{Gestalt}. While the features extracted from well segmented structures will form a Gestalt cluster, features from poorly segmented structures will fall outside this cluster. 

\begin{figure}[t] 
\begin{center}
\includegraphics[width=1.0\columnwidth]{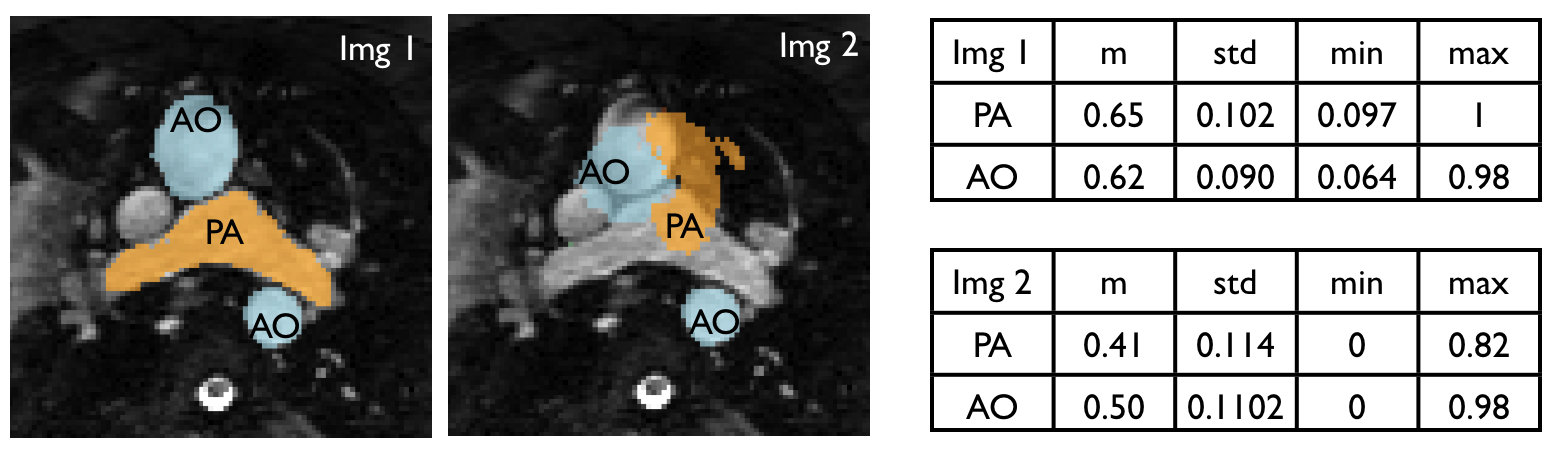} 
\end{center}
\caption{Segmentations of the aorta (AO) and the pulmonary artery (PA) are superposed to the original target image for feature extraction inside the masks. Mean (m), standard deviation (std), minimum (min) and maximum (max) values of the intensities are extracted in the example.} 
\label{fig:features}
\end{figure}

Based on this definition, we do not consider the whole image domain  for feature extraction. Instead, we superpose the segmentation obtained through MASP onto the target image. Visually significant features (\textit{e.g.} mean, standard deviation, number of connected components, max, min, etc.) capable of measuring the Gestalt grouping laws, are extracted from the masked regions (Fig.~\ref{fig:features}). 

As our aim is to diagnose d-TGA, we extract a set of 15 features from measurements obtained from the pulmonary artery and the aorta. Feature extraction is limited to these two anatomical structures as these are the only ones prone to show segmentation errors when a healthy atlas database is used. 

\textbf{Classification.}
While there are only a limited number of similar 'good' segmentations, the possible poor segmentations are many and varied. Though it is reasonable to expect well segmented images to be compactly distributed, we do not know ahead of time whether poor segmentations will be distributed in a compact or sparse and scattered way. In the former case, discriminative success will be best achieved with classical supervised learning techniques that try to best define the respective areas of dominance for the two distributions. In the latter, greater success may be achieved using unsupervised learning techniques that do not attempt to model the distribution of the scattered (\textit{negative}) class. Rather, they are recognised by their distribution as 'outliers' with respect to the compactly distributed \textit{positive} class. 

Without prior information as to which scenario we face, we apply both supervised and unsupervised methods based on the support vector machine algorithm (SVM). Briefly, in the training phase, the supervised, two class SVM attempts to find the maximal margin hyperplane that separates the two classes in an explicit feature space or an implicit one defined through the use of a kernel. 
The unsupervised, one class SVM tries to find the smallest possible hypersphere that contains a given fraction of the probability mass of a distribution in the same explicit or implicit space. 
In the classification phase, previously unseen points are assigned a binary value based on which side of the hyperplane they fall, or whether they lie inside or outside the hypersphere. 

In our context, the one-class SVM is trained using features from correctly segmented images belonging to the \textit{positive}, or suspected inlier class, and the two class SVM  is trained on features and labels from both. Input features are extracted from the images segmented using MASP. If the unseen image is classified as \textit{positive} by the two class SVM, or inlier by the one class, this implies that it is well segmented and morphologically similar to the atlas set $A$ used for its segmentation. Therefore, it can be established that the unseen image is likely to share the same clinical condition as the atlas set $A$. If the unseen image is classified as an outlier or as \textit{negative}, this suggests that it contains segmentation errors produced by the lack of morphological similarity with the atlas set $A$. In this case, it can only be established that the unseen image is unlikely to share the clinical condition as the images in $A$.

\begin{figure}[t] 
\begin{center}
\includegraphics[width=1.0\columnwidth]{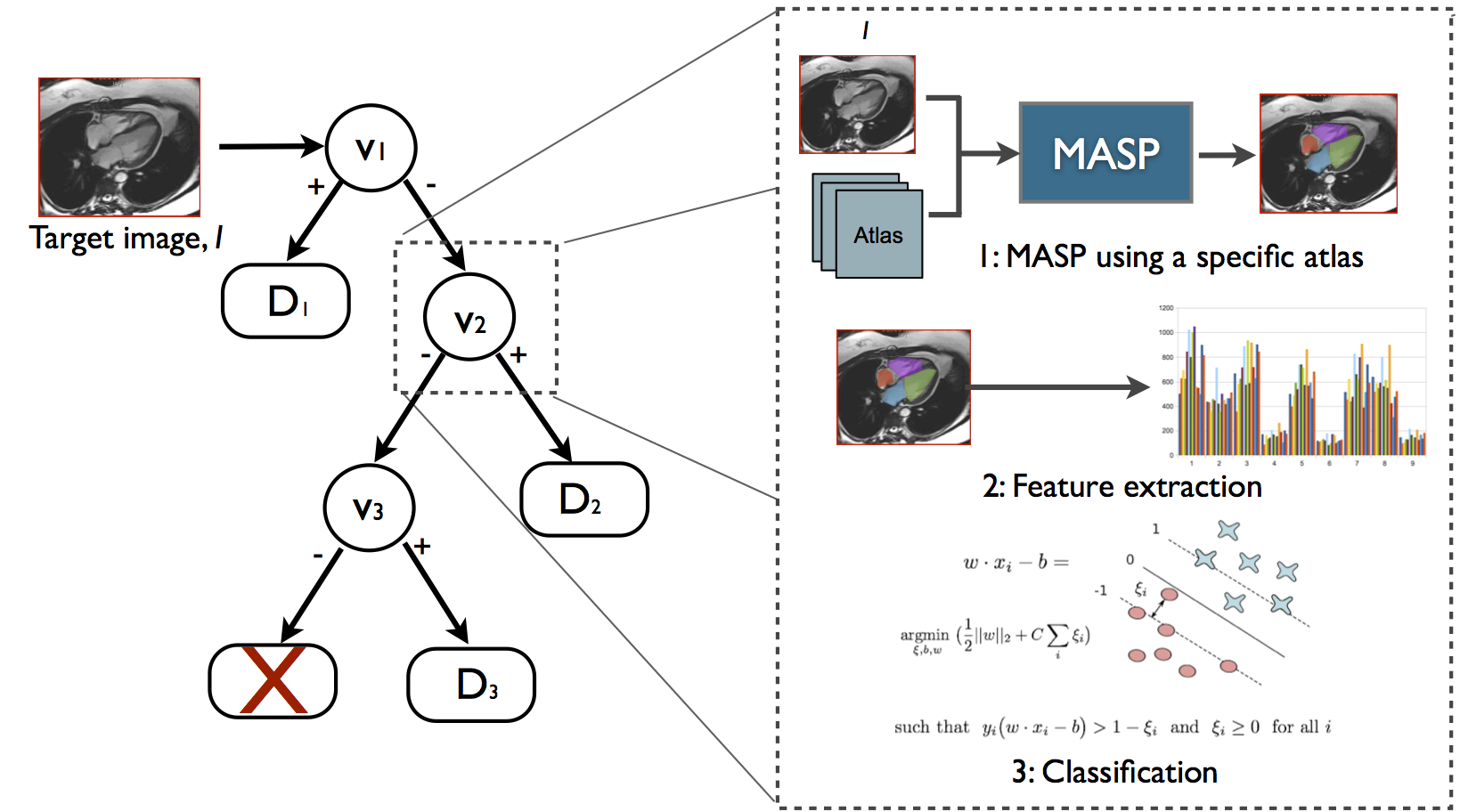} 
\end{center}
\caption{Proposed CAD system framework using a three-level logical decision tree capable of diagnosing three different conditions $D_{1-3}$ or suggesting idiopathic or uncertain diagnosis ($\times$). \textit{Positive} vs. \textit{negative} discrimination is highlighted for the second level node.} 
\label{fig:framework}
\end{figure}

\subsection{Logical Decision Tree}
For every existing pathological condition with an associated atlas set $A$ it is possible to define a decision node $v$ encapsulating the \textit{positive} vs. \textit{negative} discrimination process to determine if an unseen image $I$ is morphologically similar to $A$. Given $i \in  \lbrace 1,2,..N\rbrace$ an integer indexing the $i^{th}$ pathological condition from a vector of size $N$,  the node $v_{i}$ is going to determine if image $I$ shares the same pathological condition of the atlas set $A_{i}$.

In order to identify the underlying pathology of an unseen image $I$, a binary decision tree is built by hierarchically connecting the ensemble of decision nodes $V=\lbrace v_{1}, v_{2},...,v_{N}\rbrace$. The unseen image $I$ is positioned at the higher level decision node, $v_{1}$, for evaluation. If the \textit{positive} vs. \textit{negative} discrimination provides a positive answer, node $v_{1}$ will provide diagnose $D_{1}$, \textit{i.e.} $I$ shares the morphological characteristics of the images in $A_{1}$. If the discrimination gives a negative answer, the image will descend to the next level for evaluation. The process will be repeated until a diagnosis is given (Fig.~\ref{fig:framework}). 

The classification of $N$ different pathological conditions requires  $N-1$ decision nodes. If the final decision node $v_{N-1}$ gives a negative answer, it can be assumed the image $I$ is diagnosed with $D_{N}$. Alternatively, as depicted in Fig.~\ref{fig:framework},  an additional decision node $v_{N}$ can be included. If the node gives a negative answer, a diagnostic error will be produced to indicate that the image $I$ could not be associated to any of the pathologies represented in the tree, and requires human rater interaction.

\section{EXPERIMENTS AND RESULTS}

\subsection{Image Data}
We used a total of 60 3D steady-state free precession whole heart MRI scans acquired on a 1.5 T MR scanner (Avanto, Siemens Medical Solutions, Erlangen, Germany) at Great Ormond Street Hospital (London,UK)
for evaluation. A clinical expert classified the scans as normal controls (20), arterial switch (20) and atrial switch (20). For each clinical condition, an additional set of 5 annotated images, acquired at the same center, and containing labels of the four main chambers, the myocardium, the aorta and the pulmonary artery  was available. The three sets of images were used as atlases.

\subsection{Experiments}
The 60 MR images were segmented through MASP using the three different sets of atlases. For the classification stage, three nodes were defined: $v_{1}$ discriminates normal subjects, $v_{2}$ discriminates arterial switch and $v_{3}$ discriminates between atrial switch and a diagnosis error. The two- and one-class SVM classifiers of the nodes used a linear kernel. Tuning of the two-class and one-class SVM parameters was performed by grid search. Though the one class SVM was trained on only the inlier or positive class, both the outlier and inlier classes were used to tune its parameters in a nested cross-validation. Performance of the discriminative power of each node was validated through a leave one out strategy. A three level decision tree was constructed using the previously defined nodes and the overall system performance was evaluated using the classifier that gave the best results when assessed individually. 

\begin{figure}[t] 
\begin{center}
\begin{tabular}{c}
\includegraphics[width=0.9\columnwidth]{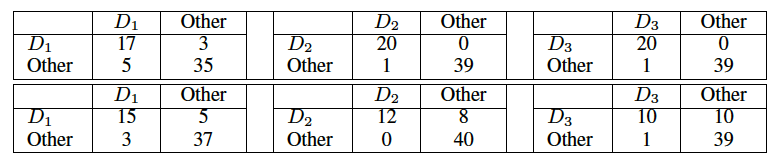} \\
\includegraphics[width=0.9\columnwidth]{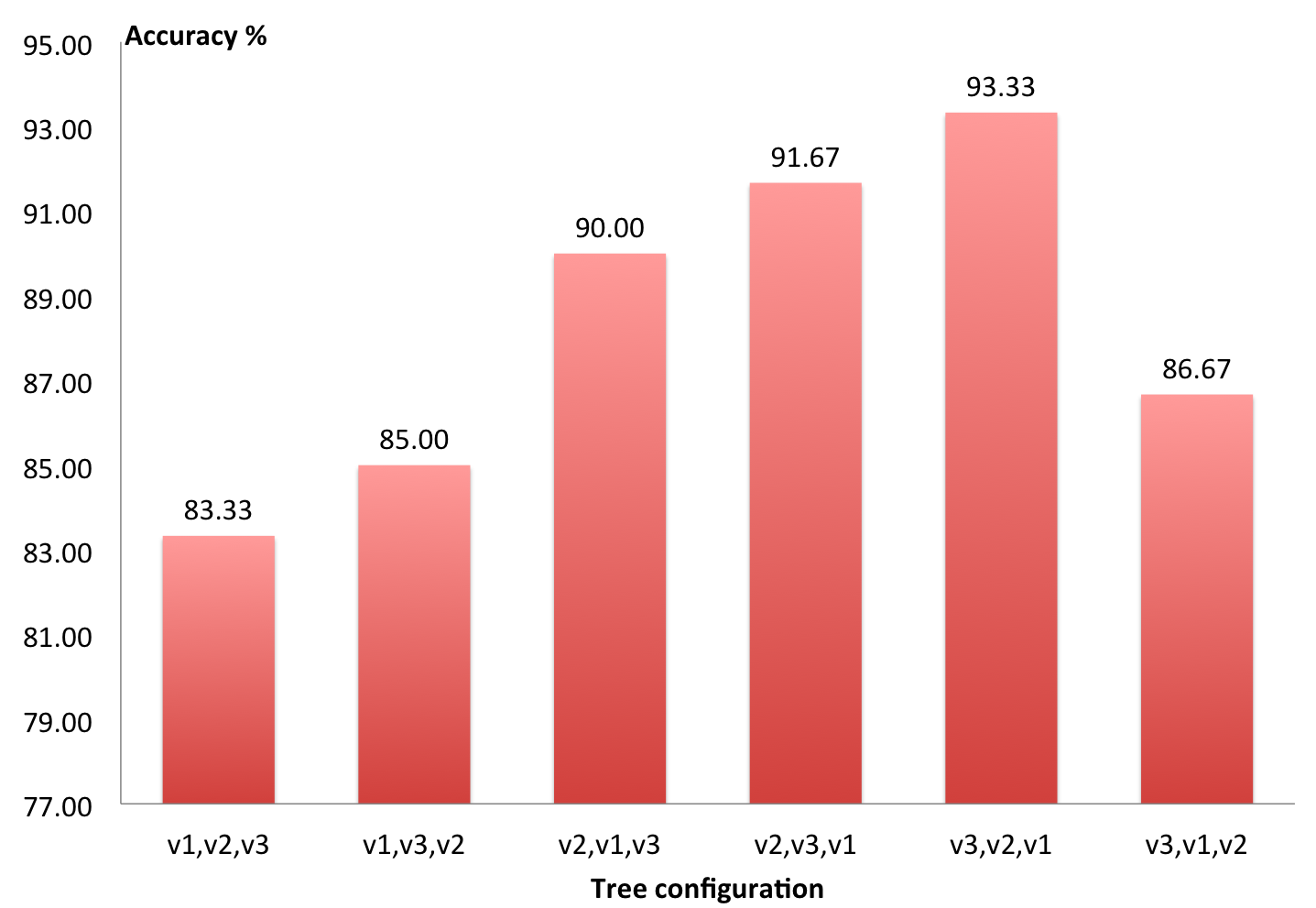} 
\end{tabular}

\end{center}
\caption{\textit{Top:} Confusion matrix for each node using  one-class SVM. $D_{1}$: normal controls, $D_{2}$: Arterial switch and $D{3}$: Atrial condition. \textit{Bottom:} Overall system accuracy as a function of the node configuration.} 
\label{fig:accuracy}
\end{figure}

\subsection{Results}
Figure~\ref{fig:accuracy} (top) presents the confusion matrices of each decision node using two- and one-class SVM, respectively. Results show that two-class SVM performs better than the one-class SVM, implying that both \textit{positive} and \textit{negative} classes have a compact distribution. The good discriminative power of the nodes is confirmed with 98.33\% accuracies for nodes $v_{2,3}$ and 86.66\% for $v_{1}$.

Using the two-class SVM as a classifier, the performance of the three-level logical decision tree was assessed. Figure~\ref{fig:accuracy} (bottom) reports the overall system accuracy as a function of the node configuration. The best overall performance (93.33\%) was achieved when $v_{3}$ was placed as the head node and $v_{1}$ was placed at the bottom.  When $v_{1}$, the node with the lowest performance is placed as head node the lowest accuracies are reported. The results show that the head node drives the CAD system accuracy.

\section{DISCUSSION AND CONCLUSIONS}
We propose a novel CAD system for pathological classification of d-TGA congenital heart disease showing an overall system accuracy of 93.33\%  in the discrimination among d-TGA variations. One of the strengths of this method is its generic nature. Although it has been applied to the identification of d-TGA, the principle can be applied for the diagnosis of pathologies of different organs. Further cardiac pathologies can be diagnosed by introducing additional decision nodes into the logical decision tree. The sole limitation for this extension is the requirement of an atlas set of annotated images describing the condition. The independence of each decision node gives large flexibility to the tree as different set of features can be used on each node. As an example, a node capable of identifying tetralogy of Fallot should focus on measuring segmentation errors in the right ventricle rather than on the great vessels.

Another interesting characteristic about the independence of the decision nodes comes from the fact that SVM provides a continuous output that is thresholded to give a final decision. This function can be tuned, independently for each node, to be more or less conservative. A conservative system increases the number of false negative (FN) outputs (uncertain diagnosis) relying mainly on an external human rater, whereas a confident system gives a high number of false positives (FP). The possibility of balancing between  FN/FP outputs is highly desirable in clinical applications.

Establishing a comparison of the proposed approach with previously published methods is not feasible as there are no previous works using atlas selection for  pathological stratification. Model selection problems within cardiac applications  have tackled other type pathologies~\cite{kutra2012} and using different sets of images that are not publicly available.  Nevertheless, the results presented in Fig.~\ref{fig:accuracy} and the overall system accuracy demonstrate the robust performance of our method.  

Among possible performance limiting factors are the segmentation quality of the 'correct' atlas set and the feature set extraction. Future work will focus on quantifying the effects of these two factors, improving the results achieved with MASP, and evaluating and selecting a larger set of features. Finally, we seek to validate  the proposed approach on a larger population to determine its potential in clinical use.  

%
%

\section*{\bf ACKNOWLEDGEMENTS}
\begin{small} 
MAZ and MJC are funded by an EPSRC grant (EP/H046410/1).
AM is funded by an EPSRC Doctoral Training Grant (EP/J500331/1) and UCL (code ELCX).  AMT is funded by the UK National Institute of Health Research (SRF/08/01/018).  SO receives funding from the EPSRC (EP/H046410/1, EP/J020990/1, EP/K005278), the MRC (MR/J01107X/1), the EU-FP7 project VPH-DARE@IT (FP7-ICT-2011-9-601055), the NIHR Biomedical Research Unit (Dementia) at UCL and the National Institute for Health Research University College London Hospitals Biomedical Research Centre (NIHR BRC UCLH/UCL High Impact Initiative).
\end{small}


\begin{thebibliography}{6}
\begin{small}

\bibitem{duchateau}
N. Duchateau, M. De Craene, G. Piella, E. Silva, A. Doltra, M. Sitges, B.H. Bijnens, A.F. Frangi: A spatiotemporal statistical atlas of motion for the quantification of abnormal myocardial tissue velocities, Medical Image Analysis 15(3), pp. 316--328 (2011)

\bibitem{zuluaga2011}
M.A. Zuluaga, D. Hush, E.J.F. Delgado Leyton, M. Hern{\'a}ndez Hoyos, M. Orkisz:
Learning from Only Positive and Unlabeled Data to Detect Lesions in Vascular {CT} Images. In: MICCAI 2011, Part III. LNCS, vol. 6893, pp. 9--16 (2011) 

\bibitem{cardoso2012}
M.J. Cardoso, K. Leung, M. Modat, S. Keihaninejad, D. Cash, J. Barnes, N.C. Fox, S. Ourselin, the Alzheimer's Disease Neuroimaging Initiative: 
STEPS: Similarity and Truth Estimation for Propagated Segmentations and its application to hippocampal segmentation and brain parcelation. Medical Image Analysis 17(6),671--684 (2013)

\bibitem{zuluaga2013}
M.A. Zuluaga, M.J. Cardoso, M. Modat, S. Ourselin:
Multi-Atlas Propagation Whole Heart Segmentation from MRI and CTA Using a Local Normalised Correlation Coefficient Criterion. In:
FIMH 2013, LNCS, vol 7945, pp. 174--181 (2013) 

\bibitem{desolneux2003}
A. Desolneux, L. Moisan, J.-M. More:
A grouping principle and four applications. 
IEEE Trans on Pattern Analysis and Machine Intelligence 25(4), 508--513 (2003)

\bibitem{kutra2012}
D. Kutra, A. Saalbach, H. Lehmann, A. Groth, S.P. Dries, M.W. Krueger, O. D\"{o}ssel, J. Weese: Automatic multi-model-based segmentation of the left atrium in cardiac {MRI} scans. In: MICCAI 2012, Part II. LNCS, vol. 7510, pp. 1--8 (2012)
\end{small}
\end{thebibliography}
\end{document}